\documentclass[letterpaper, 10pt, journal, twoside]{IEEEtran}  

\IEEEoverridecommandlockouts                              

\usepackage[sortcites,citestyle=numeric-comp,bibstyle=ieee,sorting=none,natbib=true,maxcitenames=1,mincitenames=1]{biblatex}

\usepackage{multicol}
\usepackage{hyperref}
\usepackage{booktabs}   
\usepackage[dvipsnames]{xcolor}
\usepackage[normalem]{ulem}
\usepackage{amsmath}
\usepackage{amssymb}

\newcommand{\fig}[1]  {Fig.~\ref{fig:#1}}		
\newcommand{\tab}[1]  {Table~\ref{tab:#1}}		
\newcommand{\secn}[1] {Section~\ref{secn:#1}}	

\newcommand{\Rand}{\color{red}{\texttt{\textbf{R\,}}}}
\newcommand{\Sim}{\color{red}{\texttt{\textbf{S\,}}}}
\newcommand{\Corn}{\color{orange}{\texttt{\textbf{C\,}}}}
\newcommand{\Jac}{\color{orange}{\texttt{\textbf{J\,}}}} 
\newcommand{\YCB}{\color{ForestGreen}{\texttt{\textbf{Y\,}}}} 
\newcommand{\ACRV}{\color{ForestGreen}{\texttt{\textbf{A\,}}}} 
\newcommand{\Adv}{\color{ForestGreen}{\texttt{\textbf{D\,}}}}
\newcommand{\SP}{\color{white}{\texttt{\textbf{-\,}}}}

\usepackage{graphicx}
\graphicspath{ {images/} }

\title{
EGAD! an Evolved Grasping Analysis Dataset for diversity and reproducibility in robotic manipulation
}

\author{Douglas Morrison$^{1}$, Peter Corke$^{1}$ and J\"urgen Leitner$^{1,2}$%
\thanks{Manuscript received: February 24, 2020; Accepted April, 15, 2020.}
\thanks{This paper was recommended for publication by Editor Hong Liu upon evaluation of the Associate Editor and Reviewers' comments.}
\thanks{This research was conducted under the Australian Research Council project number CE140100016, and supported by the QUT Centre for Robotics.}
\thanks{$^{1}$DM, PC and JL are with the Australian Centre for Robotic Vision (ACRV), Queensland University of Technology (QUT), Brisbane, Australia {\tt\footnotesize douglas.morrison@hdr.qut.edu.au}}
\thanks{$^{2}$JL is with LYRO Robotics, Brisbane, Australia}
\thanks{Digital Object Identifier (DOI): see top of this page.}
}

\begin{document}

\maketitle

\markboth{IEEE Robotics and Automation Letters. Preprint Version. Accepted April, 2020}
{Morrison \MakeLowercase{\textit{et al.}}: EGAD!}

\begin{abstract}

We present the Evolved Grasping Analysis Dataset (EGAD), comprising over 2000 generated objects aimed at training and evaluating robotic visual grasp detection algorithms.
The objects in EGAD are geometrically diverse, filling a space ranging from simple to complex shapes and from easy to difficult to grasp, compared to other datasets for robotic grasping, which may be limited in size or contain only a small number of object classes.  
Additionally, we specify a set of 49 diverse 3D-printable evaluation objects to encourage reproducible testing of robotic grasping systems across a range of complexity and difficulty.  
The dataset, code and videos can be found at \url{https://dougsm.github.io/egad/}

\end{abstract}


\section{Introduction}

\IEEEPARstart{T}{he} ability to grasp previously unseen objects is a fundamental trait for robots that need to interact with their environments, and underpins many higher-level manipulation capabilities.  
The last few years have seen a large amount of work focused on visual grasp detection, greatly driven by advanced deep learning techniques.  
As such, the need for diverse object dataset specific to robotic grasping is crucial for both training and evaluating these systems.  

The need for large and diverse datasets for training robust deep learning algorithms that generalise well to unknown conditions is widely recognised.  However, many current visual grasp detection algorithms are trained on either very small, manually collected datasets, or datasets of objects adapted from other domains with a small number of semantic classes, which may not be representative of the type of challenges faced in robotic grasping. 

Furthermore, there currently exists very little standardisation between the physical objects used for evaluating robotic grasping algorithms.  
While some physical object datasets do exist, they have not been widely adopted by the robotic grasping community.  Instead, researchers tend to test their algorithms using sets of random ``household'' objects, relying largely on the author's intuition as to the diversity and complexity of the test set, making effective comparison difficult.
For evaluation, results are typically reported as the average grasp success rate over an object set, however this does not allow for easy identification of system limitations or performance of a function of object difficulty or complexity.

\begin{figure}[t]
  \centering
  \includegraphics[width=1.0\columnwidth]{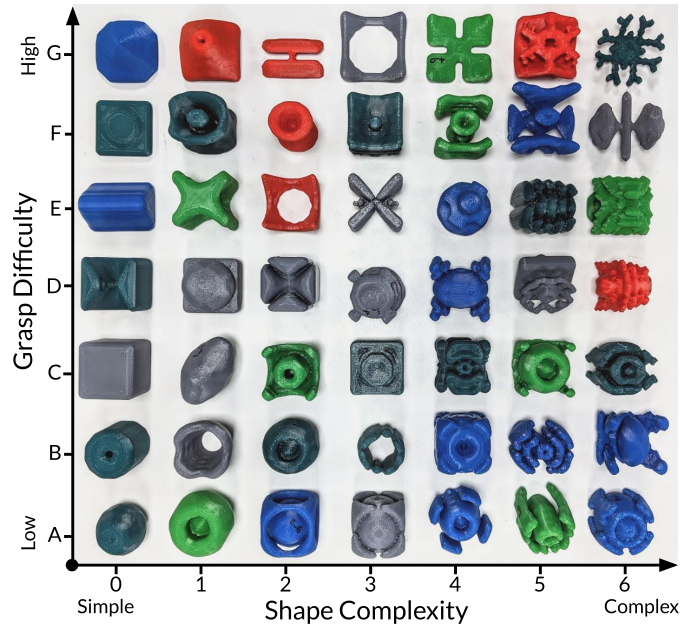}
  \caption{49 3D-printed evaluation objects chosen from over 2000 diverse objects in EGAD.  The objects provide a range of objects from simple to complex geometry (left to right), and easy to difficult graspability (bottom to top), allowing for improved and reproducible evaluation of robotic grasping algorithms.}
  \label{fig:hero}
\end{figure}

To address these issues, we use evolutionary algorithms to generate a dataset of objects that is diverse in the space of shape complexity, grasp difficulty and geometric similarity, aimed specifically at training and evaluating visual grasp detection algorithms.  The Evolved Grasping Analysis Dataset (EGAD) contains over 2000 generated objects, including a specified 3D-printable subset of 49 evaluation objects (\fig{hero}). As such, EGAD can be used for training and evaluating robotic grasping algorithms in both simulation and the real world.  To summarise our contributions, in this paper we:

\begin{itemize}
    \item Use evolutionary algorithms to create EGAD, a large dataset of over 2000 diverse objects, which fill a space of both shape complexity and grasp difficulty;
    \item Release EGAD as both 3D mesh files and in the commonly used Dex-Net~\cite{mahler2017dex} database format, with over 1 million precomputed grasp poses and the ability to easily create vision-based datasets for training grasp prediction networks;
    \item Specify a set of 49 diverse objects from the dataset which can be used as a reproducible, real-world testing suite, along with guidelines for reproducing objects and reporting results; and
    \item Perform robotic experiments using a state-of-the-art grasp detection algorithm as a template for using the evaluation set, and use the results to gain insights for future improvement of the algorithm.
\end{itemize}

\section{Related Work}

\subsection{Visual Grasp Detection}

Within the larger scope of robotic manipulation, visual grasp detection algorithms are used to predict the poses of stable robotic grasps on previously unseen objects using visual information.  Many different approaches to visual grasp detection have been proposed in recent years, with the majority using deep learning techniques~\cite{lenz2015deep, redmon2015realtime, pinto2016supersizing, johns2016deep, wang2016robot, mahler2017dex, ten2017grasp, mahler2017binpicking, kumra2017grasp, viereck2017learning, tobin2017domain, zhou2018fully, depierre2018jacquard, chu2018real, morrison2018closing, morrison2019multi, morrison2019learning, mahler2019learning, satish2019policy, asif2019densely, yu2019grasping, liang2019pointnetgpd}. 
Such approaches provide two key challenges.  The first is the availability of diverse and high quality data, which is widely recognised as an important prerequisite for training robust and generalisable models ~\cite{eysenbach2018diversity, sun2017revisiting}.  The second is providing standardised, reproducible and comparable evaluation methods and metrics.  

In other domains, large, curated datasets exist to accelerate research and standardise evaluation, e.g. ImageNet~\cite{deng2009imagenet} or COCO~\cite{lin2014coco} for computer vision.
However, the physical nature of robotics makes benchmarking and comparing work in robotic manipulation particularly difficult, with results easily influenced by robotic hardware and choice of evaluation objects. 
\textcite{mahler2018guest} provide a set of best practices for robotic grasping to ensure fair evaluation and comparison across all system aspects.  In this work we focus specifically on the aspect of object sets.

\subsection{Object Datasets for Robotic Manipulation}

The YCB object set~\cite{calli2015ycb} comprises a set of common household objects along with high-resolution 3D scans, aimed at evaluating a number of common robotic grasping and high-level robotic manipulation tasks.  However, the physical nature of this dataset means it is limited in size and object diversity from a training point of view.

On the other hand, virtual datasets for robotic grasping have been created by repurposing existing databases of 3D object meshes.  \textcite{goldfeder2009columbia} created the Columbia Grasping Dataset using 1800 3D meshes from the Princeton Shape Benchmark (PSB)~\cite{shilane2004princeton}.  Similarly, \textcite{mahler2017dex} compiled the Dex-Net dataset from 1500 3D meshes from 3DNet~\cite{wohlkinger20123dnet} and the KIT object database~\cite{kasper2012kit}.  Such datasets have proven vital for training machine-learning based grasp detection algorithms.  However, the underlying object sets are derived from 3D object recognition tasks which contain only small number of semantic classes (10 classes for 3DNet), resulting in low geometric diversity within the data sets.  

Rather than rely on realistic object models, \textcite{tobin2017domain} show that simulated objects, generated by randomly combining convex shape primitives, can be successfully used to train a grasping algorithm that generalises to real-world objects.  We build on this promising result by creating a dataset of objects that are geometrically diverse and provide a gradient of grasping difficultly and shape complexity.  

The Cornell Grasping Dataset~\cite{jiang2011efficient} provides 885 top-down RGB-D images of single objects placed on a table, hand-labelled with positive and negative grasp examples represented by a rectangle.  Due to the manual collection process, the dataset is limited in size, containing only approximately 8000 labelled grasps.  The Jacquard dataset~\cite{depierre2018jacquard} overcomes this limitation by using a simulator to generate 54k images of 11k objects, labelled with over 1 million grasps using the rectangle representation.  However, as these datasets are image-based and don't provide 3D models of objects, they are limited to training for top-down, tabletop grasping.  

\begin{table}[tpb]
    \caption{Survey of use of evaluation datasets in visual grasp detection literature. The majority of work uses irreproducible ``household'' objects for evaluation.}
    \label{tab:graspingpapers}
    \centering
    \begin{tabular}{@{}lrr@{}}
        \toprule
        \textbf{Reference} & \textbf{Evaluation Objects} &  \\
        \midrule
        \textcite{lenz2015deep}, 2015 & \Corn\SP\Rand & \SP \\
        \textcite{redmon2015realtime}, 2015 & \Corn\SP\SP \\
        \textcite{pinto2016supersizing}, 2016 & \Rand \\
        \textcite{johns2016deep}, 2016 & \Sim\Rand \\
        \textcite{wang2016robot}, 2016 & \Corn\SP\Rand \\
        \textcite{mahler2017dex}, 2017 & \Adv\SP\SP\SP\Corn\SP\Rand \\
        \textcite{ten2017grasp}, 2017 & \Rand \\
        \textcite{mahler2017binpicking}, 2017 & \Rand \\
        \textcite{kumra2017grasp}, 2017 & \Corn\SP\SP \\
        \textcite{viereck2017learning}, 2017 & \Sim\Rand \\
        \textcite{tobin2017domain}, 2017 & \YCB\SP\SP\SP\SP \\
        \textcite{zhou2018fully}, 2018 & \Corn\SP\SP \\
        \textcite{depierre2018jacquard}, 2018 & \Jac\Corn\SP\Rand \\
        \textcite{chu2018real}, 2018 & \Corn\SP\Rand \\
        \textcite{morrison2018closing}, 2018 & \Adv\ACRV\YCB\SP\SP\SP\SP \\
        \textcite{morrison2019learning}, 2019 & \Adv\ACRV\YCB\Jac\Corn\SP\Rand \\
        \textcite{morrison2019multi}, 2019 & \Adv\SP\SP\SP\SP\SP\Rand \\
        \textcite{mahler2019learning}, 2019 & \Rand \\
        \textcite{satish2019policy}, 2019 & \Adv\SP\SP\SP\SP\Sim\Rand \\
        \textcite{asif2019densely}, 2019 & \Corn\SP\Rand \\
        \textcite{yu2019grasping}, 2019 & \Rand \\
        \textcite{liang2019pointnetgpd}, 2019 & \YCB\SP\SP\SP\Rand \\

        \midrule
        \midrule
        \textbf{Legend (See text for full details)} & \\
        \midrule
        Random (``Household'') Objects & \Rand \\ 
        Objects in Simulation & \Sim\SP\\
        Cornell Grasping Dataset (IoU Metric) \cite{jiang2011efficient} & \Corn\SP\SP \\
        Jacquard SGTs \cite{depierre2018jacquard} & \Jac\SP\SP\SP \\
        YCB Objects (or subset) \cite{calli2015ycb} & \YCB\SP\SP\SP\SP \\
        APB Objects (or subset) \cite{leitner2017acrv} & \ACRV\SP\SP\SP\SP\SP \\
        Dex-Net Adversarial Objects \cite{mahler2017dex} & \Adv\SP\SP\SP\SP\SP\SP \\

        \bottomrule
        \end{tabular}
\end{table}

\begin{figure*}[tp]
  \centering
  \includegraphics[width=1.0\textwidth]{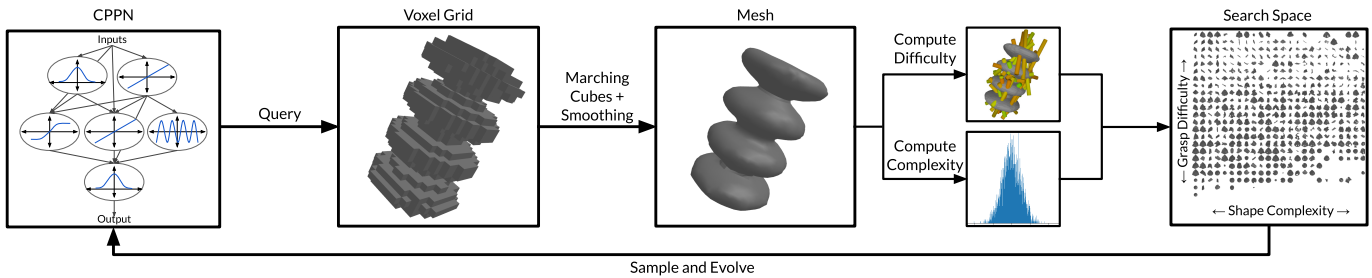}
  \caption{Overview of our method for creating EGAD.  We define a discretised search space for objects in terms of shape complexity and grasp difficulty.  Shapes are encoded using 3D CPPNs, which are queried to generate a voxel grid and processed into a 3D mesh.  We compute the shape complexity and grasp difficulty metrics to place the object in a cell of the search space.  Only the most geometrically diverse objects are kept at each cell.  At each iteration, objects are are sampled from the search space and evolved to create new objects, until the search space is full of diverse objects.}
  \label{fig:method}
\end{figure*}

\subsection{Evaluation for Robotic Grasping}

Previous work has specified datasets of physical objects and protocols for manipulation, with notable examples being the YCB dataset~\cite{calli2015ycb} and ACRV picking benchmark~\cite{leitner2017acrv}. 
However, despite the prevalence of these datasets, \tab{graspingpapers} clearly shows that neither has become commonplace for evaluating visual grasp detection systems.  Instead, authors often opt to test on sets of random ``household'' objects,
making comparing results between different algorithms very difficult, as the decision of which items are included is ultimately left to the intuition of the researchers.

\textcite{mahler2017dex} proposed a set of eight adversarial objects with complex geometry which allow testing of algorithms under difficult conditions.  The objects are easily reproducible with 3D-printing, removing the need to purchase matching objects.  As such, the objects are also scalable relative to gripper size, allowing for fairer comparisons between robotic systems.  We expand on this idea by specifying a larger and more diverse set of reproducible objects of varying complexity and difficulty that can be used to comprehensively evaluate a visual grasp detection algorithm.  

A common method of evaluation is an offline metric using the Cornell Grasping Dataset or Jacquard dataset.  A predicted grasp is successful if it has an intersection-over-union (IoU) of greater than 25\% with and is within $30^\circ$ of a positive labelled grasp when using the grasping rectangle representation~\cite{jiang2011efficient}.  While reproducible, this metric is susceptible to a large number of false-positive and false-negative detections due to the sparse labelling of the dataset, and low requirements for considering a match.  The Jacquard dataset~\cite{depierre2018jacquard} provides a cloud-based physics simulator where results are evaluated using Simulated Grasping Trials (SGTs), however this relies on a closed-source evaluation server.  Recent work has also shown that offline performance on either dataset may not be representative of real-world performance due to the domain shift from the dataset to reality~\cite{morrison2019learning}.

\section{Dataset Generation}
\label{secn:dsgen}

To facilitate both learning and evaluation for robotic grasping, we aim to provide a large set of objects that is diverse in terms of geometry, shape complexity and grasp difficulty.  For this we use evolutionary algorithms.  Compared to gradient based methods, the benefit of evolutionary algorithms in this case is the ability to handle complex, non-linear objectives in a high-dimensional design space. Evolutionary algorithms have been widely used to generate complex 3D designs, with applications ranging from art~\cite{lehman2016creative} to engineering~\cite{preen2014towards} and robotics~\cite{collins2018towards, mouret2015illuminating}.  

To summarise our approach, we first define a discretised two-dimensional search space for objects in terms of shape complexity and grasp difficulty.  Objects are encoded using Compositional Pattern Producing Networks (CPPNs)~\cite{stanley2007compositional, clune2011evolving}, from which 3D meshes are obtained.  We compute the grasp difficulty and shape complexity of each mesh to assign it to a cell in the search space.  New objects are compared to all other objects in the search space, and only the most geometrically diverse objects in each cell are kept.  The dataset is evolved using the MAP-Elites algorithm~\cite{mouret2015illuminating}, where at each iteration, objects are sampled from the search space and evolved into new, different shapes.  This process is repeated until the search space is full of diverse objects. An overview of our approach is shown in \fig{method}, and the following sections describe each component in more detail.  

\subsection{Search Space}

Our search space is defined by two features, shape complexity and grasp difficulty, and is discretised uniformly into cells.  Given an object represented by a 3D triangular mesh, it is assigned to a cell in the search space using the shape complexity and grasp difficulty metrics defined below.  A maximum number of objects is allowed at each cell.  If the number of objects in a cell exceeds the maximum, the objects are compared to all other objects in the search space, and the least geometrically diverse object (as per the metric below) is removed.  This ensures that the search space is filled uniformly and with geometrically diverse objects.  

\subsubsection{Shape Complexity}
\label{secn:complexity}

To compute a measure of shape complexity, we use the measure of morphological complexity from \cite{auerbach2014complexity, page2003shape}. The measure is based in information theory and has been shown to also correlate well with humans' intuition about shape complexity~\cite{sukumar2008towards}.  To compute the complexity metric for a given mesh, we first compute the angular deficit $\Phi_j$ for each vertex $j$:
\begin{equation}
    \Phi_j = 2\pi - \sum_i \phi_i
\end{equation}
where $\phi_i$ is the internal angle of each triangle $i$ where it meets vertex $j$.  The deficit values are placed in a histogram over the range $[-2\pi, 2\pi)$ with bin width $\Delta$, which is normalised as a probability density function (PDF) such that each bin $b$ contains a probability $p(\Phi_b)$. The shape complexity is then equivalent to the entropy of the PDF:
\begin{equation}
    H = - \sum_b p(\Phi_b) \log p(\Phi_b)
\end{equation}

\subsubsection{Grasp Difficulty}
\label{secn:gq}

To estimate a single scalar feature representing grasping difficulty per object, we use the 75\textsuperscript{th} percentile method described by \textcite{wang2019adversarial}.  Using the Dex-Net analytical grasp planner~\cite{mahler2017dex}, we sample a number of antipodal grasps on each object and compute the robust Ferrari-Canny quality metric for each. The grasp difficulty feature is then obtained by taking the 75\textsuperscript{th} percentile grasp quality of all sampled grasps.  

\subsubsection{Geometric Diversity}
\label{secn:diversity}

In order to compute the geometric diversity of an object, we first define a metric of geometric similarity between any two objects. We use the Topology Matching metric based on Multiresolutional Reeb Graphs (MRGs) proposed by \textcite{hilaga2001topology}, which provides a shape similarity score $\text{\textit{sim}} \in [0,1]$ between two object meshes that is robust to translation, rotation, scale and changes in mesh connectivity (e.g. through mesh resampling or decimation).  This method has been shown to work effectively with arbitrary meshes and CAD models~\cite{hilaga2001topology, bespalov2003reeb}.  We define the distance \textit{dist} between two object meshes $m_1$ and $m_2$ as the inverse of similarity:
\begin{equation}
    \text{\textit{dist}}(m_1, m_2) = 1 - \text{\textit{sim}}(m_1, m_2)
\end{equation}

Similar to \cite{gomes2015devising}, we then define the diversity of a mesh $\rho(m)$ as the mean distance to the $k$ most similar meshes to $m$ in the whole search space:

\begin{equation}
    \label{eq:novelty}
    \rho(x) = \frac{1}{k} \sum_{i=1}^k \text{dist}(m, m_i)
\end{equation}

\subsection{Evolutionary Algorithm}
\subsubsection{Shape Encoding}
\label{secn:cppn}

We use Compositional Pattern Producing Networks (CPPNs)~\cite{stanley2007compositional, clune2011evolving} to encode and generate 3D shapes.  Each CPPN is an arbitrary neural network, which allows for a compact, functional representation of a 3D volume obtained by querying the network at discrete spatial coordinates (e.g. $x$, $y$ and $z$), and thresholding the scalar output to create a voxel grid.  The voxel grid is converted to a triangle mesh representation using marching cubes~\cite{lewiner2003efficient} and smoothing is applied.  In the case that multiple disconnected meshes are generated, we keep only the largest volume mesh.  Small features that would prevent 3D-printing are removed by performing a morphological opening.  

CPPNs are evolved by the principles of NeuroEvolution of Augmenting Topologies (NEAT)~\cite{stanley2002evolving} using the NEAT-Python library~\cite{neat-python}.  At each evolution, the CPPN architecture is mutated by randomly adding and removing network nodes and connections, and changing weights, biases and activation functions.  CPPNs also undergo crossover, where components of two CPPN architectures are combined into a new CPPN.  As such, the CPPNs and their resulting shapes build up complexity and diversity over time.  

\subsubsection{Search Algorithm}

Many traditional evolutionary algorithms are very effective optimisation algorithms, but are susceptible finding local minima and not exploring the search space.  To overcome this, we use the Multi-dimensional Archive of Phenotypic Elites (MAP-Elites) algorithm~\cite{mouret2015illuminating}.

Our implementation of MAP-Elites begins with a population of randomly initialised CPPNs, which are queried and placed into their respective cells of the search space.  At each subsequent iteration, a population is randomly sampled from the search space to undergo evolution and produce a new population of objects, which are subsequently evaluated and assigned to cells in the search space.  If any cell contains more than the maximum number of individuals, the lowest performing (least geometrically diverse) occupants are iteratively removed until all cells contain at most the maximum number of occupants.  The diversity of all occupants is recalculated after each removal.

\section{Evolved Grasping Analysis Dataset}

\begin{table}[tpb]
    \caption{Parameters used for dataset generation}
    \label{tab:exp1_params}
    \centering
    \begin{tabular}{@{}lr@{}}
        \toprule
        \textbf{Parameter}  & \textbf{Value} \\
        \midrule
        Population Size & 100 \\
        Evolution Steps & 200,000 (2000 steps $\times$ population 100) \\
        Search Space Size & 25$\times$25 \\
        Max Objects per Cell & 4 \\
        Probability of Crossover & 0.5 \\
        Difficulty Feature Range & [0.0005, 0.004] \\
        Complexity Feature Range & [1, 5] \\
        Histogram Bin Width ($\Delta$) & $4\pi / 512$ \\
        Sampled Grasps per Object & 100 \\
        Diversity neighbours ($k$) & 10 \\
        CPPN Resolution & $25\times25\times25$ \\
        CPPN Inputs & $x$, $y$, $z$, $\sqrt{x^2 + y^2}$, $\sqrt{x^2 + z^2}$, \\
        \multicolumn{2}{r}{$\sqrt{y^2 + z^2}$, $\sqrt{x^2+y^2+z^2}$} \\
        \multicolumn{2}{r}{$x, y, z\in[-1,1]$} \\
        CPPN Activation Functions & sin, sigmoid, gaussian, idenity \\
        \bottomrule
        \end{tabular}
\end{table}

\begin{figure*}[tp]
  \centering
  \includegraphics[width=1.0\textwidth]{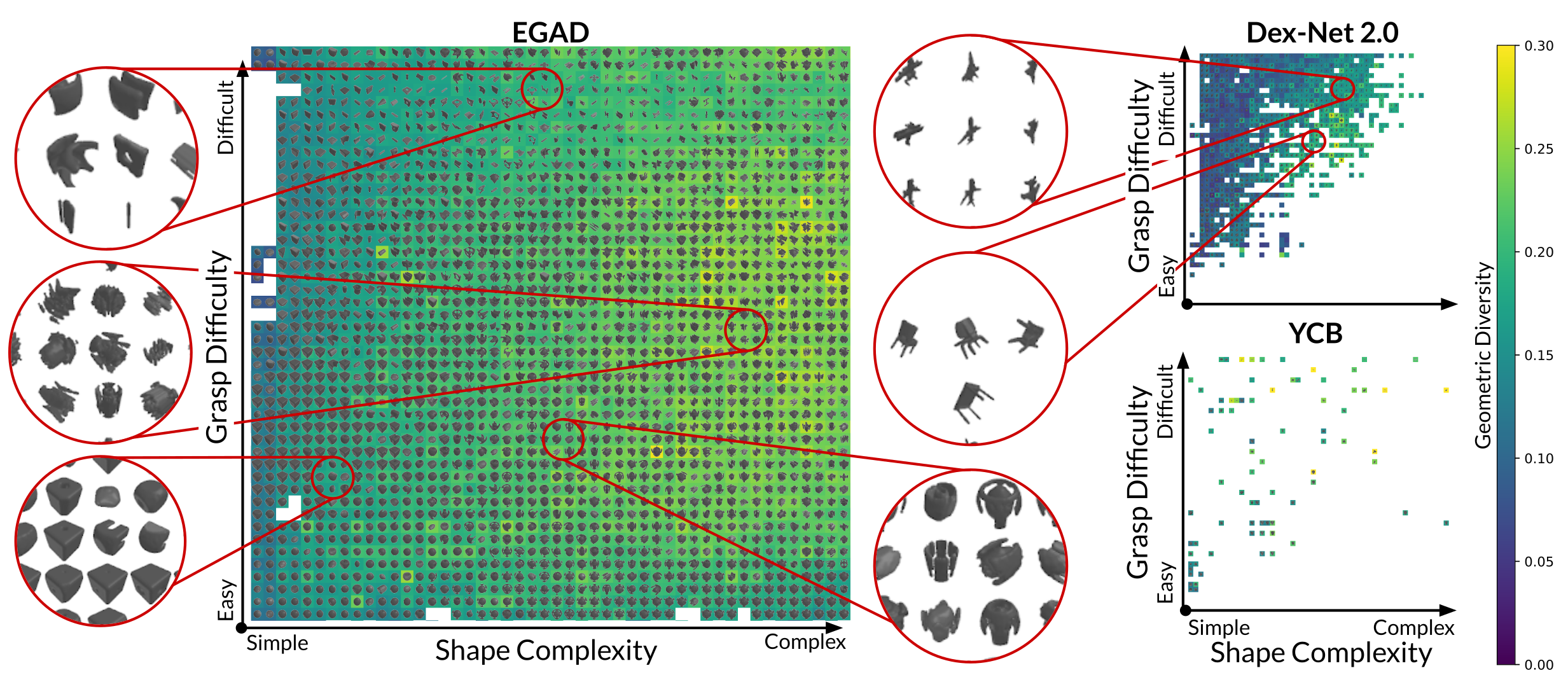}
  \caption{The distribution and diversity of EGAD in our object search space, compared to the object models found in Dex-Net 2.0 and YCB object sets.  NB: For YCB objects, only models with an associated laser scan were used.  High resolution versions are available on the \href{https://dougsm.github.io/egad/}{project webpage}.}
  \label{fig:egad}
\end{figure*}

Using the method described in \secn{dsgen} and the parameters defined in \tab{exp1_params}, we generate a set of objects which we call the Evolved Grasping Analysis Dataset (EGAD).  In this section we present the dataset with analysis of the objects and a comparison to other existing grasping datasets.  

\subsection{Overview}

In total EGAD contains 2331 objects which uniformly fill the search space.  While the total size of the search space is 2500 ($25\times25\text{ cells}\times4\text{ objects per cell}$), some cells at the extremes were unable to be filled, e.g. objects that are in the simplest geometric column but difficult to grasp, or geometrically complex and easy to grasp. The coverage of EGAD is thus 93\% of the search space.  The distribution of objects and their diversity is shown in \fig{egad}.

The diversity of the objects ranges from 0.07 to 0.27, with a mean of 0.19.  Unsurprisingly, the lowest diversity objects largely found in the area of simple shape complexity, and the object diversity increases with higher shape complexity.

Many of the objects exhibit symmetry about multiples axes, which is a result of the CPPN inputs which provide distances to the $x$, $y$ and $z$ planes and origin in addition to the $x$, $y$ and $z$ positions.  This is advantageous, as it makes the grasping difficulty of the objects less sensitive to changes in orientation.

\begin{table}[tpb]
    \caption{Comparison of EGAD with other grasping datasets}
    \label{tab:comparison}
    \centering
    \begin{tabular}{@{}lccccc@{}}
        \toprule
        & & & \multicolumn{3}{c}{\textbf{Diversity}} \\
        \cmidrule{4-6}
        \textbf{Dataset} & \textbf{Size} & \textbf{Coverage} & \textbf{Min} & \textbf{Max} & \textbf{Mean}  \\
        \midrule
        YCB~\cite{calli2015ycb} & 78 & 0.03 & 0.11 & 0.65 & 0.19 \\
        Dex-Net 2.0~\cite{mahler2017dex} & 1497 & 0.37 & 0.03 & 0.26 & 0.11 \\
        EGAD & 2331 & 0.93 & 0.07 & 0.27 & 0.19 \\
        \bottomrule
        \end{tabular}
\end{table}

\subsection{Dataset Comparison}

For comparison, we also embed the YCB and Dex-Net 2.0 object datasets into the same search space (\fig{egad}), with a quantitative comparison also given in \tab{comparison}.  Due to the small number of objects, the YCB has only very sparse coverage of the object search space, filling 3\% of all cells.  The small number of objects leads to an overall high object diversity mean of 0.19.  However, this is still equivalent to EGAD which contains orders of magnitude more objects.  

The Dex-Net 2.0 object dataset is much larger than YCB, but still only covers 37\% of the space, and also exhibits much lower geometric diversity than EGAD on average.  This is due to the small number of semantic classes of the object models, many of which are geometrically simple shapes (e.g. the container, fruits and foods categories from 3DNet), and similarly, the more complex parts of the space are largely filled by very similar instances of aircraft and chairs.  

\begin{figure}[tp]
  \centering
  \includegraphics[width=0.85\columnwidth]{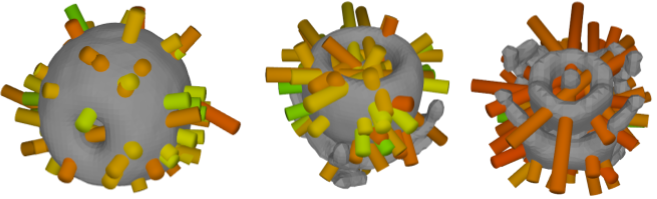}
  \caption{Examples from the Dex-Net compatible database of EGAD objects with pre-computed grasps. Colours indicate the robust grasp quality metric, ranging from red to green.}
  \label{fig:dex}
\end{figure}

\subsection{Mesh Dataset} 

We release EGAD in two formats.  The first format is a set of 3D triangular meshes, which can be easily adapted to many robotics tasks such as training grasping algorithms or simulation.  Compared to other available mesh datasets (e.g.~\cite{wohlkinger20123dnet, shilane2004princeton}), EGAD provides a number of benefits, in particular: (1) the dataset has been created specifically for robotic manipulation, so it provides a diverse set of objects which are applicable to the task, rather than meshes representing only a small number of semantic classes; (2) objects are labelled according to their complexity and difficulty, and provide a range in both of these dimensions; (3) the generated meshes are watertight, cohesive (i.e. no missing faces, inverted normals, etc.) and don't contain unnecessary internal detail, making direct usage in simulation, rendering/visualisation and reality easier (not the case for other datasets~\cite{wohlkinger20123dnet, mahler2017dex}); and (4) meshes have been post-processed to allow 3D-printing, making sim-to-real comparisons possible.

To aid adoption of EGAD, the second format leverages the widely used Dex-Net project~\cite{mahler2017dex} to create a Dex-Net compatible database of EGAD objects.  Each object is pre-labelled with up to 100 antipodal grasps labelled with a robust grasp-quality metric (\fig{dex}).  The Dex-Net project provides extra functionality, including the ability to label objects with other grasp sampling strategies or quality metrics, add custom grippers, and generate large image datasets for training visual grasp detection algorithms.

In addition, we provide code to enable the creation of custom object datasets for specific applications.  
While EGAD is specific to antipodal grasps, other gripper types, e.g. multifinger, can be accommodated by replacing the grasp sampling method in \secn{gq}.  Similarly, domain or task specific object distributions can be created by imposing geometric constraints on the CPPN output.

\subsection{Evaluation Set}

As robotic grasping is an inherently physical problem, we believe that testing on a physical robotic system is the most important step in evaluating such systems.  As such, we designate a set of 49 objects from EGAD as an evaluation set, which can be 3D-printed and used to test real-world robotic systems  (\fig{hero}).  49 objects serves as a practical number for manual robotic testing while also providing a good spread of difficulty and complexity.
The objects were chosen in an automated manner from the full dataset such that they are uniformly representative of the object space in a $7\times7$ grid while minimising geometric similarity between objects within the evaluation set.  For evaluation, these objects provide a gradient of complexity and difficulty, which can identify the strengths and limitations of visual grasp detection algorithms better than collections of commonly used items.

Objects are labelled alphanumerically according to their position in a $7\times7$ grid (\fig{hero}) with A0 being the simplest and easiest object, and G6 being the most complex and difficult.

\begin{figure}[tp]
  \centering
  \includegraphics[width=1.0\columnwidth]{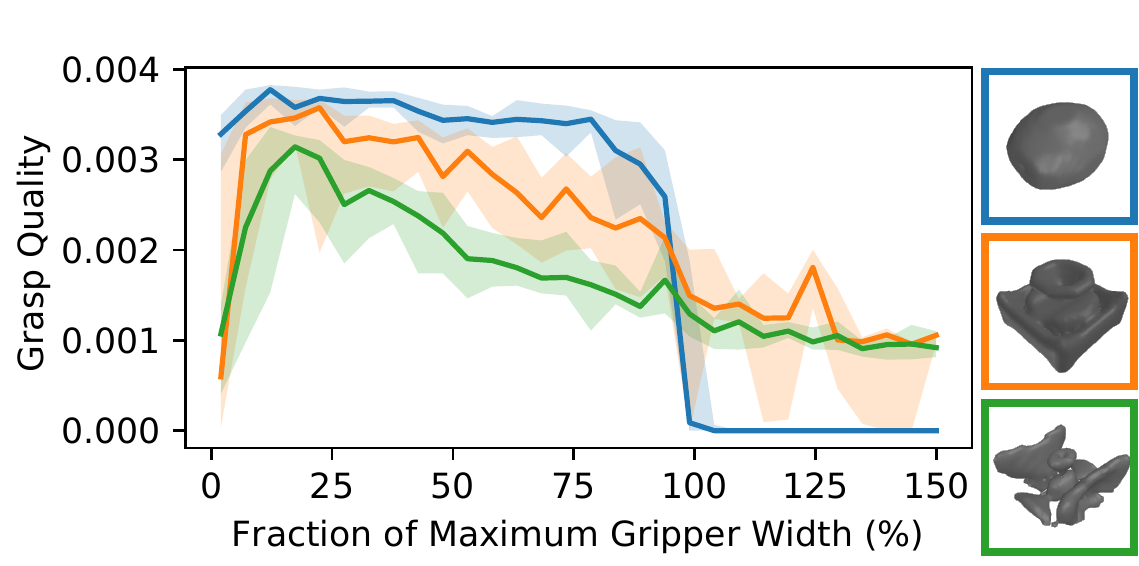}
  \caption{75\textsuperscript{th} percentile grasp quality (\secn{gq}) as a function of object size as a fraction of maximum gripper width for three objects, highlighting the importance of object size when reproducing grasping experiments. Shaded area shows 65-85\textsuperscript{th} percentile.}
  \label{fig:size}
\end{figure}

\subsection{Effect of Object Size}

One factor that effects the grasping difficulty of an object is its size relative to the gripper.  \fig{size} illustrates this on three different objects from EGAD, highlighting that the size effects the overall graspability in a nonlinear way depending on the object's geometry.  Compared to 3D printing, a downside to datasets comprising commercial products is that grasp performance may be largely effected by the choice of robotic gripper, making isolation and direct comparison of the visual grasp prediction aspect more difficult.

To ensure uniformity in our results and to aid in reproducibility, we use a constant scaling factor throughout all of our experiments, for both dataset generation and in producing the physical evaluation set.  Each object is scaled such that its minimum bounding box dimension is no more than 80\% of the gripper width. Along with our dataset, we also release a Python script for re-scaling the provided meshes by this metric for a given gripper width to allow fair results comparisons.

\section{Robotic Experiments}

In the following sections we use the EGAD evaluation set to evaluate a state-of-the-art visual grasp detection network.  In doing so we provide a template for presenting results.  The diversity of the evaluation set allows us to gain insights into the strengths and limitations of the visual grasp detection approach, providing future research directions to improve upon these baseline results.  

\subsection{Visual Grasp Detection}

We use the Generative Grasping Convolutional Neural Network (GG-CNN) from \cite{morrison2018closing,morrison2019learning} as a visual grasp detection algorithm.  GG-CNN is a fully convolutional network that provides a one-to-one mapping from an input depth image to a prediction of grasp quality and pose at every pixel in realtime.  Each pixel in the output defines a top-down grasp pose, defined by $\mathbf{g} = (\mathbf{c}, \phi, w, q)$, where $\mathbf{c} = (x,y,z)$ is the position of the grasp's centre, $\phi$ is a rotation around the vertical axis, $w$ is the desired width of the gripper and $q$ represents a grasp quality.  GG-CNN was trained on the Cornell Grasping Dataset~\cite{jiang2011efficient}.

\subsection{Equipment}

For robotic grasping experiments we use a Franka Emika Panda robot, fitted with 3D-printed fingers with silicone tips based on \cite{guo2017design}.  The maximum opening of the fingers is 75mm, and the evaluation objects were scaled according to this.  An Intel Realsense D435 depth camera is attached to the end effector of the robot to provide visual input.  Videos of the experiments are available on the \href{https://dougsm.github.io/egad/}{project webpage}.

\subsection{Procedure}

We perform 20 grasp attempts on each of the 49 3D-printed evaluation objects in isolation, for a total of 980 grasp attempts.  Objects were placed one at a time into the workspace of the robot.  For each grasp attempt, a depth image was captured from a fixed viewpoint and the best predicted grasp from GG-CNN was executed by the robot.  The grasp success rate is the fraction of grasps after which the robot successfully lifted and held the object 40cm above the table.  If the grasp was successful, the object is dropped into the workspace to randomise the pose and position for the next grasp attempt.  Objects that fall outside of the workspace are replaced.  

\begin{figure}[tp]
  \centering
  \includegraphics[width=0.75\columnwidth]{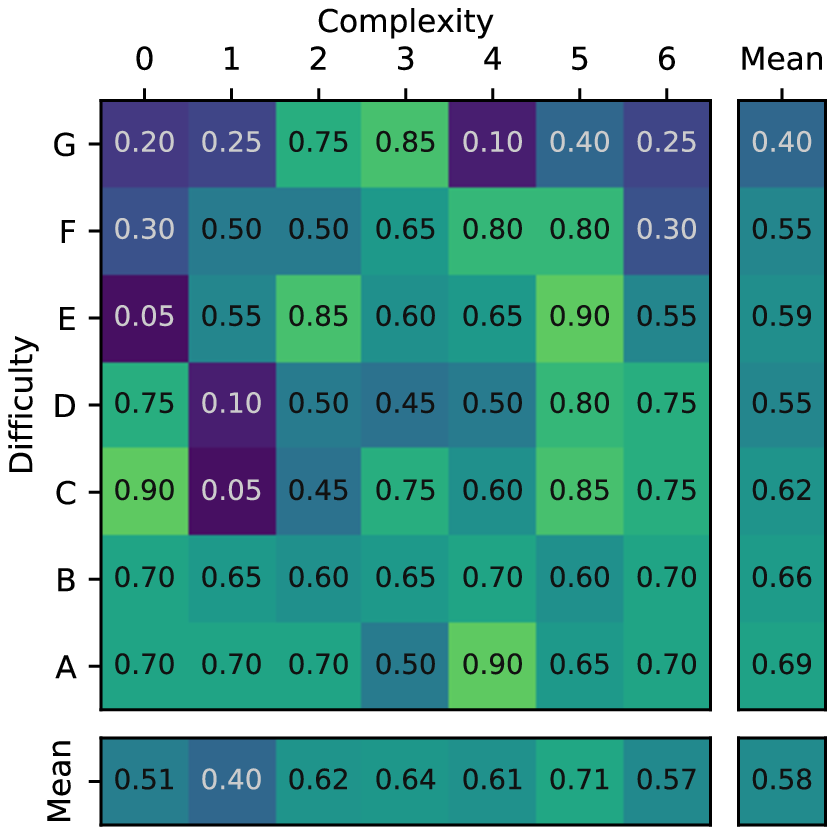}
  \caption{Average grasp success rate for each object in the evaluation set.  Labels correspond to the evaluation objects as shown in \fig{hero}. Outer cells show the mean over the respective axes.}
  \label{fig:results}
\end{figure}

\subsection{Results and Discussion}
\label{secn:discussion}

\fig{results} shows the grasp success rate for each of the 49 3D-printed evaluation objects individually.  The right-hand column shows that the grasp success rate decreased from 69\% for the easiest 7 objects (A row), to 40\% for the 7 most difficult (G row).  The overall success rate across all objects was 58\%.  While this is lower than previously reported results for GG-CNN~\cite{morrison2018closing, morrison2019multi, morrison2019learning}, this is expected since the EGAD objects are more complex and difficult than those previously used.

Despite the trend in difficulty, there are some clear outliers.  For example, objects C1 and D1 performed much worse than their neighbours, and objects E5, F4, F5 and G3 that performed much better despite being difficult objects.  
Using the EGAD evaluation objects in this way allows us to analyse these results in a more introspective way than reporting a single success rate on a non-diverse object set.  
In the rest of this section we discuss some identified strengths and limitations of GG-CNN with regards to these results, and propose future improvements.

\begin{figure}[tp]
  \centering
  \includegraphics[width=1.0\columnwidth]{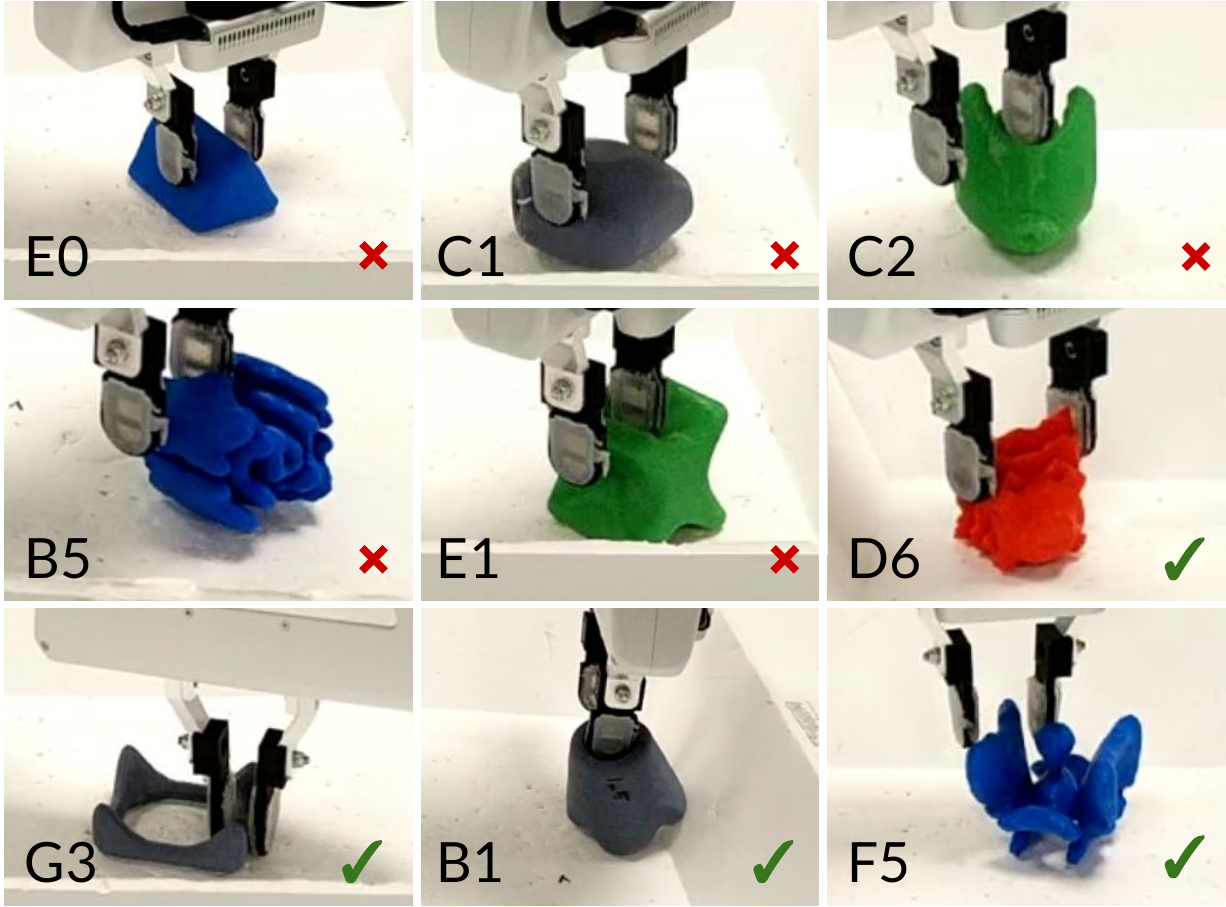}
  \caption{Examples chosen from experimental grasp attempts to highlight strenghts and limitations of GG-CNN.  {\color{red}{$\mathbf{\times}$}} indicates grasp failure and {\color{ForestGreen}{\textbf{\checkmark}}} indicates success.  Refer to \secn{discussion} for details.}
  \label{fig:analysis}
\end{figure}

\textbf{Grasp Depth} A number of simpler and easier objects, in particular E0, C1 (pictured in \fig{analysis}) and D1, have very low grasp success rates.  This highlights a major failure case for GG-CNN, where the depth of the grasp is computed relative to the grasp's centre, resulting in grasps that are too shallow resulting in failure.  Round objects in the range A0 to B2 were also effected in the same way.
A major improvement to GG-CNN may be to also encode the required depth of the grasp in the input or prediction.

\textbf{Orientation Bias}  The grasps generated by GG-CNN are often oriented perpendicular to an object's major axis.  For example, on object E0 this results in a high number of failures due to the sloping sides (\fig{analysis}). Meanwhile, grasping the object lengthways would provide a stable grasp on the object's flat ends.  This bias is likely caused by the lack of similar examples in the training data, in which grasps on long objects are heavily biased towards perpendicular grasps, and could be corrected with improved training data.

\textbf{Top-Down Grasping}  Like many other visual grasp detection algorithms, GG-CNN is limited to producing top-down (4-DoF) grasps.  However, for objects such as C2, there are limited ways to grasp the object from a top-down orientation (\fig{analysis}).  This provides a strong motivation for using 6-DoF grasps that align to the graspable parts of objects.

\textbf{Grasping Object Parts}  Unlike many other visual grasp detection algorithms, GG-CNN predicts the gripper width for each grasp.  This is advantageous for objects such as B1 (in certain orientations), G3 and F5 (\fig{analysis}), where precise, narrow grasps are required to avoid collision with object parts, resulting in a higher than average success rate despite these objects' complexity and difficulty.  On the other hand, a number of failures were noted for objects such as B5 and E1, where depressions in the objects cause grasps that result in collisions and grasp failures.  

\textbf{Finger Material} In addition to visual aspects, physical properties of the gripper also effect the results. Objects such as E5 and D6 (\fig{analysis}) are largely difficult due to their uneven surface with many acute protrusions, making grasps performed with a rigid gripper surface unstable.  However, the compliant nature of the silicone fingertips used, as explored by~\cite{guo2017design}, largely accounts for this by moulding to the surface resulting in stable grasps and a high success rate for these objects.

\section{Conclusion}

We presented EGAD, a dataset of over 2000 evolved 3D objects for training and evaluating robotic grasping and manipulation.  The objects uniformly fill a space of shape complexity and grasp difficulty, compared to other similar datasets which are limited in both size and diversity.  This provides the necessary diversity for training robust visual grasp detection algorithms.  Additionally, we specify a diverse evaluation set of 49 objects which are 3D-printable to allow for reproducible testing of grasping algorithms over a wide range of complexity and difficulty.  

Using the EGAD evaluation set, we were able to identify a number of limitations of a state-of-the-art grasping algorithm GG-CNN, which has previously not been possible on simpler sets of ``household'' objects.  In future work we propose to use these insights to improve on the baseline results, and to investigate the effect of diverse training data on the robustness of visual grasp detection algorithms. 

\section{Acknowledgement}

We would like to thank David Howard for his insight and technical guidance in designing the evolutionary algorithm, Akansel Cosgun for his constructive feedback throughout the project, and Jack Collins for his advice on algorithmic details and parallelisation for HPC.




\printbibliography

\end{document}